%% file: paper.tex
\documentclass[journal]{vgtc}                
\ifpdf
  \pdfoutput=1\relax                   
  \usepackage{graphicx}                
  \DeclareGraphicsExtensions{.pdf,.png,.jpg,.jpeg} 
\else
  \usepackage{graphicx}                
  \DeclareGraphicsExtensions{.eps}     
\fi%

\graphicspath{{figures/}{pictures/}{images/}{./}} 
\usepackage{CJKutf8}
\usepackage{microtype}                 
\PassOptionsToPackage{warn}{textcomp}  
\usepackage{textcomp}                  
\usepackage{mathptmx}                  
\usepackage{times}                     
\usepackage{cite}                      
\usepackage{tabu}                      
\usepackage{booktabs}                  
\usepackage{graphicx}
\usepackage{amsmath}
\usepackage[utf8]{inputenc}
\usepackage{ifthen}
\usepackage{soul, color, xcolor}

\definecolor{ColorNameTODO}{RGB}{0,0,0}
\definecolor{ColorNameRevised}{RGB}{0,0,0}
\definecolor{ColorDelete}{RGB}{255,35,25}

\newcommand{\Revised}[1]{{\color{ColorNameRevised}#1}}
\newcommand{\Del}[1]{{\color{ColorDelete}}}

\onlineid{1066}

\vgtccategory{Research, Application}
\vgtcpapertype{system}

\title{C\textsuperscript{5}: Towards Better Conversation Comprehension and Contextual Continuity for ChatGPT}

\author{Pan Liang, Danwei Ye, Zihao Zhu,Yunchao Wang, Wang Xia, Ronghua Liang, \textit{Senior Member, IEEE}, and Guodao Sun*}
\authorfooter{
\item
Pan Liang, Danwei Ye, Zihao Zhu, Yunchao Wang, Wang Xia, Ronghua Liang, and Guodao Sun are with College of Computer Science and Technology
College of Software, Zhejiang University of Technology. Guodao Sun* is the corresponding author. E-mail: guodao@zjut.edu.cn.
}

\shortauthortitle{Biv \MakeLowercase{\textit{et al.}}: C5: Towards Better Conversation Comprehension and Contextual Continuity for ChatGPT}

\abstract{Large language models (LLMs), such as ChatGPT, have demonstrated outstanding performance in various fields, particularly in natural language understanding and generation tasks. In complex application scenarios, users tend to engage in multi-turn conversations with ChatGPT to keep contextual information and obtain comprehensive responses. However, human forgetting and model contextual forgetting remain prominent issues in multi-turn conversation scenarios, which challenge the users' conversation comprehension and contextual continuity for ChatGPT. To address these challenges, we propose an interactive conversation visualization system called C\textsuperscript{5}, which includes Global View, Topic View, and Context-associated Q\&A View. The Global View uses the GitLog diagram metaphor to represent the conversation structure, presenting the trend of conversation evolution and supporting the exploration of locally salient features. The Topic View is designed to display all the question and answer nodes and their relationships within a topic using the structure of a knowledge graph, thereby display the relevance and evolution of conversations. The Context-associated Q\&A View consists of three linked views, which allow users to explore individual conversations deeply while providing specific contextual information when posing questions. The usefulness and effectiveness of C\textsuperscript{5} were evaluated through a case study and a user study.
} 


\CCScatlist{ 
 \CCScat{K.6.1}{Management of Computing and Information Systems}%
{Project and People Management}{Life Cycle};
 \CCScat{K.7.m}{The Computing Profession}{Miscellaneous}{Ethics}
}

\teaser{
  \centering
  \includegraphics[width=\linewidth]{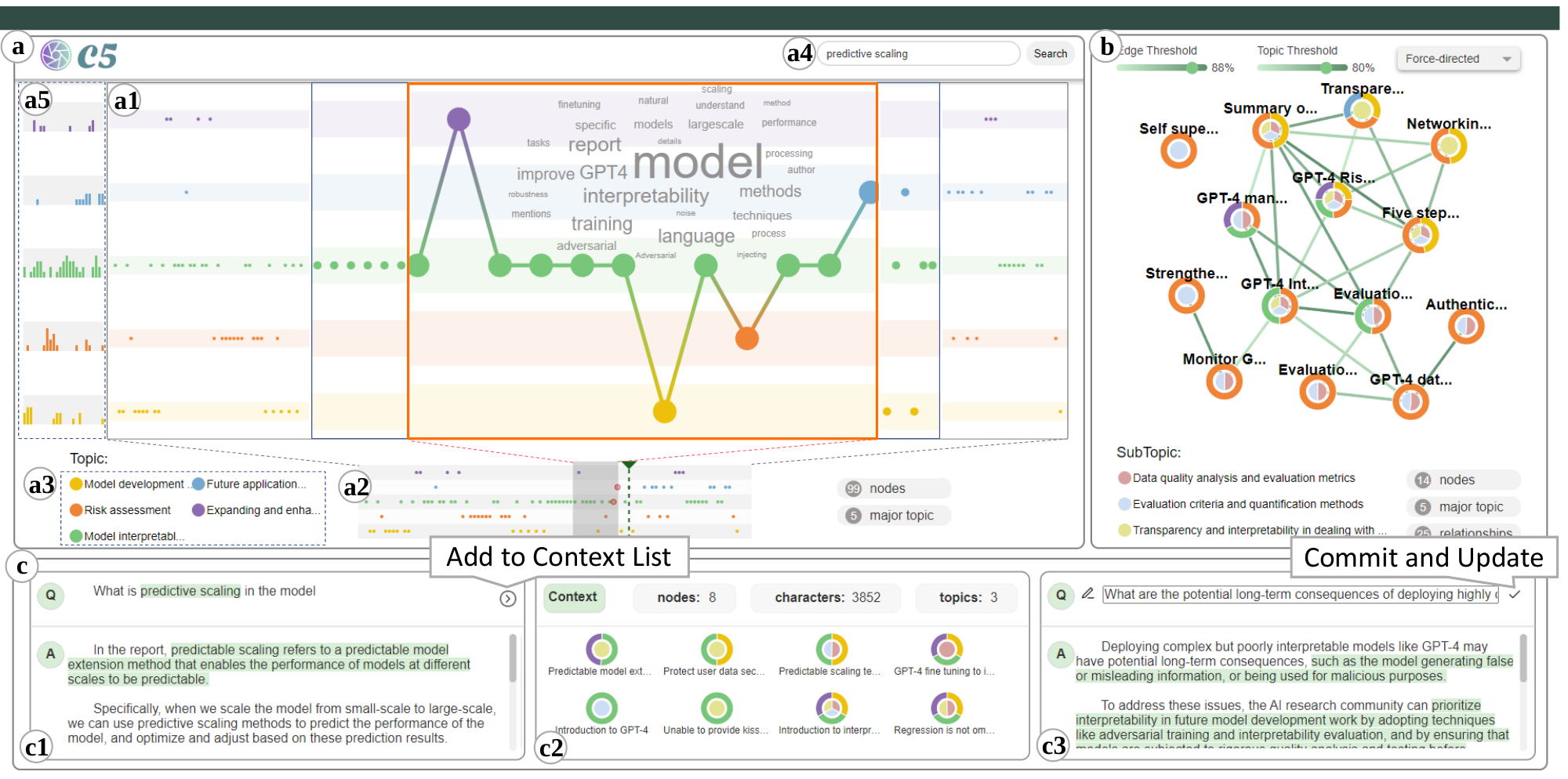}
  \caption{User interface: (a) Global View for presenting the trend of conversation evolution and supporting the exploration of locally salient features. (b) Topic View for displaying the relevance and evolution of conversations within a topic. (c) Context-associated Q\&A View for allowing users to explore individual conversations deeply while providing specific contextual information when posing questions.}
  \vspace{-0.5em}
	\label{fig:teaser}
}

\vgtcinsertpkg

\begin{document}


\input{sections/Introduction}

\input{sections/relatedwork}

\input{sections/System_Overview}

\input{sections/C5}

\input{sections/Evaluation}
\input{sections/DISCUSSION_AND_LIMITATIONS}
\input{sections/Conclusion}
\input{sections/ACKNOWLEDGMENT}

 \bibliographystyle{plain} 
 \bibliography{paper}

\end{document}

%% file: sections/Introduction.tex
\firstsection{Introduction}

\maketitle
In recent years, Large Language Models (LLMs) have made significant progress and gained wide application in various fields\cite{liu2023summary,jiao2023chatgpt,2023arXiv230308774O}. ChatGPT, an advanced LLM, leverages its large database and efficient design to understand and interpret user requests, generating accurate answers for complex scenarios\cite{liu2023summary,jiao2023chatgpt,2023arXiv230308774O}. In complex application scenarios, users tend to engage in multi-turn conversations with ChatGPT to maintain contextual information and obtain comprehensive responses. This interaction pattern allows the model to better understand user intent, subsequently providing more precise responses. 


\Revised{\textbf{Human forgetting} and \textbf{model contextual forgetting} are two significant challenges in multi-turn conversation scenarios with ChatGPT, especially when users engage with the model for complex tasks\cite{savelka2023can,yang2023exploring,lecler2023revolutionizing}. \textbf{Human forgetting} becomes evident as conversations encompass multiple rounds and various sub-topics intertwined rather than in a linear fashion. Initially, users typically start with a broad goal or task, which eventually fragments into sub-tasks or themes during the conversation. As users delve into a specific sub-topic, their focus narrows down, and upon conclusion, when attempting to return to the primary line of inquiry, they often realize the loss of grasp over prior dialogues. The convoluted, intermingled nature of the conversation threads complicates the retrospective location and retrieval of relevant segments, affecting \textbf{conversation comprehension}.

In parallel, \textbf{model contextual forgetting }stems from LLMs’ token limitations which impair their ability to maintain \textbf{contextual continuity} across a dense multi-themed conversation\cite{yang2023exploring}. When users present a query that does not directly relate to the immediately preceding discourse, the model, restricted by its linear context association, may falter in generating accurate or contextually congruent responses.

These challenges necessitate a system that facilitates the analysis of conversation history, catering to both the user's and the model's contextual limitations. Such a system should enable effective retrieval and review of relevant conversation threads, aiding users in recollecting critical information, and providing the model with the means to contextually anchor responses.}

Previous studies have focused on text visualization\cite{cui2011textflow,knittel2021real,el2016contovi,shi2018meetingvis,cowell2006understanding} and optimization of the internal structure of LLMs\cite{li2023unlocking,niu2021review,katharopoulos2020transformers,liu2023pre}. However, text visualization mainly focuses on the visual presentation of textual topic evolution, while ignoring the detailed presentation of content as well as the complete structure and complexity of the conversation, which limits the user's comprehensive grasp and in-depth analysis of the overall conversation structure. Moreover, although some progress has been made in optimizing the internal structure of LLMs, they do not fully explore the potential of user involvement in improving the contextual associations of these models.

This work aims to solve the above challenges through a visualization approach, for which we develop an interactive conversation visualization system named C\textsuperscript{5}. \Revised{Our system is primarily aimed at individuals and professionals who frequently engage with ChatGPT for complex information retrieval and decision-making processes.} The system intuitively presents the content of multi-turn conversations through simple visual elements and clear layouts to assist users in efficiently understanding and grasping the content of conversations, while assisting ChatGPT in tracing historical conversation information, ultimately enhancing conversation comprehension for users and contextual continuity for ChatGPT. We propose a topic modeling approach based on GPT, which classifies conversation histories into topics at different granularities and generates hierarchical topic-based text data with temporal information. To present complex data results and address the challenges mentioned, our visual analytics system includes three interactive views: Global View, Topic View and Context-associated Q\&A View. Global View uses the GitLog diagram metaphor to present the conversation structure, where different topics are encoded as different strips in the horizontal direction, and to reduce the wiggle of nodes between topics, we use a quadratic assignment-based optimization method to determine the vertical position of topics. This demonstrates the trend of conversation evolution and supports the exploration of locally salient features. Topic view reveals the relevance and evolution of conversations within a single topic. The Context-associated Q\&A View facilitates in-depth exploration of individual conversations, while enabling users to provide context-specific information when posing questions, thereby enhancing contextual continuity for ChatGPT. Our contributions are as follows:
\begin{itemize}
 \vspace{-0.5em}
    \item A pipeline for effectively solving the human forgetting problem and providing a new visualization-based perspective for addressing contextual forgetting in LLMs.
    \vspace{-0.5em}
    \item C\textsuperscript{5}, an interactive conversation visualization system, clearly presents the conversation content and structure. It assists users in efficiently exploring the conversation history and provides contextual information when posing questions, thereby enhancing conversation comprehension for users and contextual continuity for ChatGPT.
     \vspace{-0.5em}
    \item A comprehensive evaluation, demonstrating the effectiveness and usability of C\textsuperscript{5}. By using both a detailed case study and a user study with ChatGPT users, we establish the practical advantages of our system in resolving the challenges presented in multi-turn conversation scenarios.
     \vspace{-0.5em}
\end{itemize}

%% file: sections/relatedwork.tex
\section{Related Work}

\subsection{Text Embedding and Topic Modeling}

Text embedding is the process of converting textual data into numerical representations, which is crucial for natural language processing and machine learning tasks. In addition to traditional bag-of-words models, there are many advanced text embedding methods. Word embeddings such as Word2Vec\cite{church_2017} and GloVe\cite{pennington2014glove} capture semantic and syntactic information by training neural networks that map words to a high-dimensional continuous vector space. Sentence or paragraph embedding methods, such as Doc2Vec \cite{le2014distributed}, extend the concept of word embeddings to represent longer texts as fixed-length vectors. Recently, pre-trained context-based word embeddings, such as BERT \cite{devlin2018bert} and RoBERTa \cite{liu2019roberta}, can capture the different meanings of words in various contexts, further improving the quality of text embedding. Additionally, OpenAI has recently introduced large-scale pre-training models like GPT-3 \cite{2020arXiv200514165B}, which also provide API interfaces for direct user calls to achieve high-quality text embedding. 

Topic modeling is a technique that employs natural language processing and machine learning to discover potential topics in large amounts of text data, and is widely used in fields such as social media analysis, news reporting, and academic research. Latent Dirichlet Allocation (LDA) \cite{blei2003latent} is the most commonly used topic modeling algorithm, which estimates the topic composition of documents and the word distribution of topics using generative probability models and Bayesian inference. Non-negative Matrix Factorization (NMF) \cite{lee2000algorithms} decomposes the term-document matrix into two non-negative matrices, representing the topics and their corresponding weights in each document. Spherical k-Means \cite{dhillon2001concept} is based on the widely used k-Means clustering algorithm, utilizing cosine distance instead of Euclidean distance to improve the processing of text data. The Neural Topic Model (NTM) \cite{Cao_Li_Liu_Li_Ji_2015} models topic-word distributions and document-topic ratios using neural networks. TopicRNN \cite{2016arXiv161101702D} is a recurrent neural network that models topic changes over time. Additionally, the currently popular LLMs, such as GPT \cite{radford2018improving,liu2023summary,2023arXiv230308774O}, have also achieved good results in topic modeling, demonstrating superior performance compared to traditional topic models in various NLP tasks across multiple datasets\Revised{\cite{2023arXiv230308774O,wei2022emergent,hendy2023good,lai2023chatgpt,ahuja2023mega}}. 

After evaluating and comparing various text embedding and topic modeling methods, we found that approaches based on LLMs demonstrate significant advantages in handling natural language processing tasks. Therefore, after comprehensively considering the strengths and weaknesses of different methods, we decided to adopt the GPT-based approach for text embedding and topic modeling.

\subsection{Temporal Topic Visualization}
In recent years, temporal Topic Visualization has received considerable attention \cite{knittel2021real,wu2017streamexplorer,sun2014evoriver,kim2016topiclens}. ThemeRiver\cite{havre2002themeriver}, a classic text visualization technique, uses the river metaphor to visualize topics over time. TextFlow \cite{cui2011textflow} enables analysts to clearly explore and analyze the merging and splitting of topics through sankey and pileup diagrams relationships over time. RoseRiver\cite{cui2014hierarchical} proposed an interactive visual text analysis approach for exploring and analyzing complex evolutionary patterns of hierarchical topics using topic trees and tree cuts. Liu et al\cite{liu2015online} it proposed a tree- and sedimentation-based visualization of topics in text streams that uses EvoBRT \cite{knittel2021real} for clustering. MeetingVis\cite{shi2018meetingvis} employs visual narrative-based summarization through topic bubbles and word clouds to effectively trigger recall of subtle details and enhance productivity in team-based workplaces by summarizing meeting content.

The aforementioned approaches focus on visual exploration of evolving topics, including content change and strength change. Unlike these approaches, our work emphasizes the mantra ``overview first, zoom and filter, then details-on-demand'', progressively exploring conversation history from the overall conversation structure to the details of individual Q\&A sessions. Technically, we propose a visual analysis method that supports multi-granularity topic exploration, addressing the challenge of human forgetting by providing a visual analytics system that includes  multiple views to assist users in progressively mining and reviewing the historical content of multi-turn  conversations. Our method integrates the advantages of previous research on topic evolution-based text visualization and specifically optimizes for multi-turn conversation scenarios, thereby enhancing users' conversation comprehension.

\Revised{\subsection{Conversation Visualiztion}}

\Revised{

Previous conversation visualizations primarily focus on capturing and representing discussions among several participants. ChAT\cite{cowell2006understanding} utilizes a matrix to track interactions and topics, while Meeting Adjourned\cite{ehlen2008meeting} employs a linear timeline for topics and action items. SUVI\cite{castronovo2008generic} integrates chronological and relevance-based layouts. ConToVi\cite{el2016contovi} implements a sedimentation metaphor for conversational flows, and IdeaWall\cite{shi2017ideawall} groups content thematically. NEREx\cite{el2017nerex} focuses on relationships between named entities in transcripts. Additionally, MeetingVis\cite{shi2018meetingvis} is a visual narrative-based tool that processes spoken audio from meetings and displays summarized content, facilitating recall and reflection on participant activities, topic evolutions, and task assignments.

Contrasting with these approaches, our work centers on conversations between two parties - the user and ChatGPT. It emphasizes the logical structure and content, supporting users in a structured exploration of the conversation and comprehending its progression and content.
}

\subsection{Contextual Association  Enhancement with LLMs}
LLMs\cite{lewis2019bart,raffel2020exploring} represented by GPT \cite{floridi2020gpt} have made important progress in the field of NLP with powerful generation and understanding capabilities. However, all of these models suffer from contextual association limitations, resulting in their potential inability to adequately capture and retain historical contextual information when dealing with multi-turn conversations. 

A significant amount of research has focused on optimizing the internal structures of LLMs to address these context association limitations. For example, efficient attention mechanisms and memory-augmented networks have been employed to enhance the capture and retention of historical context information in multi-turn conversations\cite{niu2021review,katharopoulos2020transformers}. Furthermore, self-information filtering has been used to eliminate low-information content, thus enhancing the efficiency of fixed context lengths\cite{li2023unlocking}. These approaches mainly focus on optimizing the internal structures of LLMs during the training phase.

\Revised{In contrast, there are also some studies that explore how to enhance the performance of LLMs during the utilization phase. For instance, EMNLP\cite{madaan2022memory} improves the performance of GPT-3 by searching for questions with similar intent and obtaining user feedback. SELF-REFINE\cite{madaan2023self} generates initial outputs using LLMs, and the same LLM provides feedback on these outputs for iterative improvements. Entailer\cite{tafjord2022entailer} showcases how answers are derived through a series of systematic reasoning chains reflecting its internal beliefs. Additionally, LlamaIndex connects large language models with external data through a query-retrieval approach. These works explore innovative ways to harness the capabilities of LLMs during the utilization phase.

Building on these insights, our work enhances the performance of LLMs during the utilization phase by designing an interactive visual conversation analysis system. In our system, new user queries are treated as search queries, with the conversation history serving as a database. Through vectorization, relevant dialogues within the conversation history are searched, similar to LlamaIndex. Our unique contribution is the integration of visualization tools to address contextual association limitations. This not only mitigates the issue of context forgetting in LLMs but also provides a more intuitive and efficient method for users to interact with these models by focusing on context-specific information. Our approach is novel and holds practical value in scenarios where retaining contextual information is crucial for meaningful multi-turn conversations with LLMs.
}




%% file: sections/System_Overview.tex
\section{System Overview}

Before proceeding with the system design, we first focused on the collection of system design requirements. To achieve this goal, we referred to the relevant papers \cite{shi2018meetingvis, shneiderman1996eyes, el2016contovi} to understand the consensus and requirements about such systems. Subsequently, we conducted in-depth interviews and discussions with ChatGPT users and domain experts to better understand user needs and gain expert guidance on text topic evolution and conversation visualization.

We conducted a comprehensive analysis of past research and summarized three consensus requirements: providing a visual representation of the conversation history, following the mantra ``overview first, zoom and filter, then details on demand'' and ensuring the interactivity and user-friendliness of the system, enabling users to easily explore and analyze conversation data.

To further understand user needs, we interviewed 5 ChatGPT Plus users, including two doctoral students, two master's students, and one undergraduate student, who worked in the research fields of natural language processing, computer vision, and visualization. Since December of last year, they had been using ChatGPT, with current usage times ranging from 2 to 5 hours per day. These users stated that ChatGPT had high practicality in complex tasks such as paper reading, report writing, and programming. However, users encountered several problems when using ChatGPT for these complex tasks. For example, they might forget earlier conversation content, or conversation efficiency would be reduced due to ChatGPT losing contextual information. To address these issues, users expressed expectations for our system, including presenting a global overview of conversation history and visualizing the content of individual conversations. Additionally, they hoped the system would support tracking conversation history to prevent ChatGPT from losing contextual information, enabling it to provide more accurate and useful answers.

Simultaneously, we discussed the technical aspects with two visualization experts. These experts possessed extensive research and practical experience and had published in top conferences and journals in the visualization field. They provided useful insights on text topic evolution and conversation visualization. Based on these discussions, we adapted and optimized the design solution to ensure the feasibility, scalability, and robustness of the system.

By combining the results of past research, user interviews, and discussions with domain experts, we obtained a clear and comprehensive set of system design requirements. These requirements provided a solid basis for the detailed description of the system overview, design tasks, and workflow in the subsequent chapters.

\subsection{System Requirements}
Our system requirements are as follows:

\textbf{R1: Provide an intuitive representation of conversation history. }Recognizable, explorable, simplified visuals make it easy for users to understand conversation content by lowering cognitive threshold.

\textbf{R2: Organize conversation elements to trigger recall of specific memories. }Revealing conversation structure that supports memory recall by organizing related conversation information rather than simply presenting independent conversation information.


\textbf{R3: Allow ChatGPT to associate conversation content from any time.} Provides a dynamic interactive way to display conversation history and enhance the performance of ChatGPT association context through user engagement.

\subsection{Design Tasks}
Based on above requirements, we propose specific visualization tasks to guide the development of C\textsuperscript{5}, following the mantra ``summarize first, zoom and filter, then provide detailed information as needed'' \cite{shneiderman1996eyes}.

\textbf{T1: Global overview of conversation history.} The system should provide an overview of conversation history in chronological order, and show the connections and transitions between different topics, so that users can understand the relationships and influences between topics, thus helping the model establish contextual correlations (\textbf{R1}).


\Revised{\textbf{T2: Zoom and in-depth exploration.} The system should facilitate zooming into specific sections of the conversation history for a detailed analysis. This is key for understanding the intricacies and context of individual threads, which is vital for meaningful multi-turn conversations with ChatGPT (\textbf{R1}).}

\textbf{T3: Topic selection and display of conversation content.} The system should support users in clicking on specific topic, displaying all relevant conversation nodes within the topic, while revealing the degree of association and evolution, so that users can understand the depth and breadth of the topic, and establish a rich knowledge system (\textbf{R2}).

\textbf{T4: Highlighting.} The system should support highlighting in several respects: (1) The system should highlight historical conversations that are highly relevant to the user's interests. (2) The system should emphasize historical conversations that are closely related to the questions posed by the user. (3) The system should underscore key phrases within individual conversations to alleviate the burden on users of reading large blocks of text (\textbf{R2, R3}).

\textbf{T5: Detailed information and overview of Q\&A nodes.} The system should provide detailed information display for each Q\&A node, and provide an overview of the content of a single Q\&A, so that users can view more content according to their needs or quickly browse ChatGPT answers (\textbf{R3}).

\subsection{Workflow}
\begin{figure*}[!t]
  \centering
  \includegraphics[width=\textwidth]{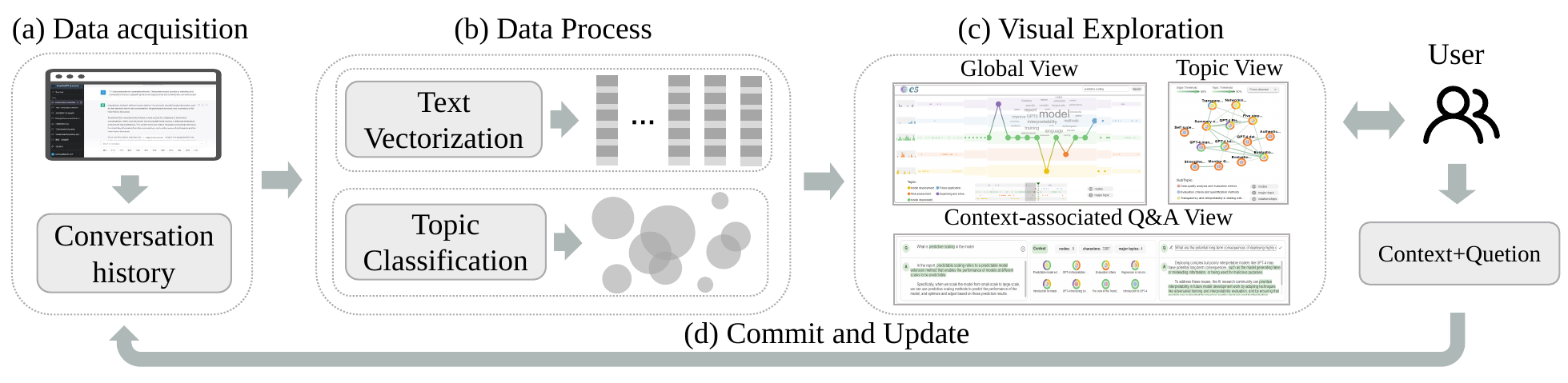}
  \caption{C\textsuperscript{5} workflow. (a) we first obtain the conversation history data from the ChatGPT web page and divide it into individual conversation nodes. (b) these data are processed using text embedding and topic classification.  (c) the user can interactively explore conversation history through three visual components that compose our framework. Finally, (d) we achieve incremental updates through real-time integration and visualization, in order to display the latest conversation history. }
  \vspace{-1.5em}
  \label{fig:workflow}
\end{figure*}
After mapping the requirements to design tasks, we defined the workflow of the system (Fig. \ref{fig:workflow}) in four key stages: data acquisition, data processing, visualization, and incremental update.

\textbf{Data Acquisition} (Fig. \ref{fig:workflow}(a)), we obtain the conversation history of users with ChatGPT and decompose this into individual conversation nodes. Every node has a specific structure and content. Specifically, each node includes a user's question and ChatGPT's corresponding response.

\textbf{Data Processing} (Fig. \ref{fig:workflow}(b)), we perform text embedding on conversation history for better handling and analysis. Additionally, we perform multi-granularity topic classification of the conversation history to reveal knowledge structures and associations for visualization presentation.

\textbf{Visualization} (Fig. \ref{fig:workflow}(c)), users can explore conversation history through a visualization interface and provide specific context information when posing questions. The visualization interface includes three interactive views: Global View (displaying conversation history and topic associations), Topic View (displaying internal associations and evolution within topics), and Context-associated Q\&A View (presenting original text and providing context selection functions).

\textbf{Incremental Update} (Fig. \ref{fig:workflow}(d)), when users pose new questions, the system helps users add specific context information and sends it along with the question to ChatGPT. Subsequently, the system automatically repeats the first three stages, integrating new content and updating the visualization interface in real-time.

Through these four stages, our workflow provides an intuitive and efficient interface (details in Sec. 4), helping users better explore conversations while providing sufficient context information for ChatGPT, thereby enhancing users' conversation comprehension and ensuring contextual continuity for ChatGPT.

%% file: sections/C5.tex
\section{Conversation Visualization}

\subsection{Data Processing}
\begin{figure}[htbp]
  \centering
  \includegraphics[width=\columnwidth]{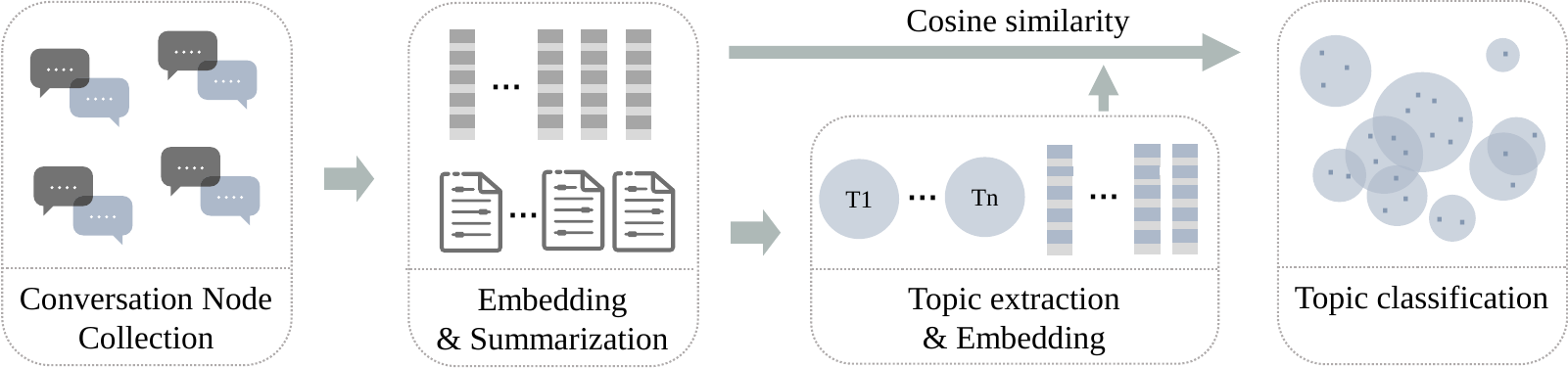}
  \caption{Data Processing. (a) conversation node generation through collecting and preprocessing conversation history records. (b) text embedding using a GPT-based model for effective representation of text information. (c) topic classification based on GPT-3.5 Turbo to generate and embed latent topics. (d) subtopic division using recursive execution of the classification.}
  \vspace{-1.5em}
  
  \label{fig:dataprocessing}
\end{figure}

In this section, we discuss the \textbf{data processing stage} (Fig. \ref{fig:dataprocessing}) in detail, which consists of three key steps: conversation node generation, text embedding, and topic extraction and classification. The goal of the data processing stage is to convert the conversation history between users and ChatGPT into hierarchical topic-based text data with temporal information.

\textbf{Conversation node generation}: First, we collect conversation history records of users' conversations with ChatGPT from web page. Then, these records are pre-processed to extract the questions and corresponding answers from them, and they are treated as a conversation node (denoted as $n$). Subsequently, we arrange all conversation nodes according to the order they were generated, forming a conversation node set with temporal information (denoted as $N$).

\textbf{Text embedding}: To achieve effective representation of text information, we employ a embedding model based on GPT-3.5. This model demonstrates strong semantic extraction capabilities and computational efficiency, making it suitable for text embedding in complex contexts. After embedding, each conversation node obtains a vector representation (denoted as $v_{n}$), and all vector representations of conversation nodes form a vector set (denoted as $V_{n}$).

\textbf{Topic extraction and classification}: We propose a topic classification method based on GPT-3.5 Turbo to overcome the limitations of traditional machine learning methods (such as LDA). The topic classification process includes the following steps: (a). To adapt to the maximum token limitation of GPT-3.5 Turbo, we summarize each conversation node, then add the summary result to $N$. (b). We construct appropriate prompts, allowing GPT-3.5 Turbo to generate a series of latent topics (denoted as $T$) based on the summaries in the $N$ , and then embed these topics to obtain the vector representations of topics (denoted as $V_{t}$). (c). We calculate the cosine similarity between the $V_{n}$ and $V_{t}$, and determine whether each conversation node belongs to a specific topic based on a predefined threshold. It is worth noting that a conversation node may belong to multiple topics. (d). For each topic internally, we recursively execute the above process to achieve subtopic division.

After the above processing steps, we obtain hierarchical topic-based text data with temporal information, revealing the latent structure of the conversation history.

\subsection{Global View}
The Global View (Fig. \ref{fig:teaser}(a)) is the main entry panel of the system, designed to provide a full view of conversations (\textbf{T1}) in a temporal perspective and to help users locate conversation history of interest through an interactive interface (\textbf{T2}). The Global View primarily consists of Content View (Fig. \ref{fig:teaser}(a1)) and Brush View (Fig. \ref{fig:teaser}(a2)) to present the data obtained through data processing.

In Brush View, users can observe the distribution of different topic in the entire conversation history from a temporal perspective (\textbf{T1}), with each topic encoded in different colors. The x-axis represents the sequence of conversations, while the y-axis represents the topics to which the conversations belong. Brush View has a ``\textbf{forgotten line}'' (the green dashed line in Brush View), the position of which is calculated based on the maximum number of token ChatGPT can process. The purpose of the ``forgotten line'' is to remind the user when posing a question that the conversation node located before the ``forgotten line'' is beyond ChatGPT's contextual association capabilities. The topic legends are displayed on the left side of Brush View (Fig. \ref{fig:teaser}(a3)). If users are interested in a particular topic during exploration, they can click the corresponding label to switch the display content of Topic View (\textbf{T3}). Furthermore, in (Fig. 1(a5)), users can clearly visualize the temporal distribution of each individual topic.

The Content View magnifies the area brushed by users in the Brush View (\textbf{T2}). Similar to Brush View, the positions of nodes are determined by order and topic, and connected by solid lines. \Revised{In complex conversation histories, relationships among dialogue components can be intricate. Solid lines offer an intuitive means to visually represent the continuity and progression of the conversation, emphasizing not only the temporal sequence but also the way topics evolve and sometimes interconnect. This visual representation is crucial for users to holistically grasp the conversation’s structure, making it easier to identify patterns, trends, and topic transitions. Furthermore, as conversations naturally flow and can contain overlapping themes, the connections support an enhanced contextual understanding.}The word cloud displays the keywords of the conversations within the brushed area\Revised{, with the size of each word representing its importance computed by using TF-IDF weighting scheme instead of using mere word frequency. This ensures that the size of a word in the word cloud is reflective not only of its frequency but also of its significance to the conversation, thus helping to bring attention to more relevant words.} When users hover over a word, the system highlights nodes based on the word's frequency in each conversation node, helping users quickly understand the main discussion topics and focal points in the area (\textbf{T4}). In addition, users can hover over nodes of interest (especially those at topic transitions) and the system will display an overview diagram of that Q\&A content (\textbf{T5}). To view the original conversation text (\textbf{T5}), users can click on the corresponding node to access detailed information in the Q\&A view (Fig. \ref{fig:teaser}(c1)). This view provides users with an effective approach to discover and explore salient local features of the conversation history for enhancing users' conversation comprehension.

Moreover, there is a search box(Fig. \ref{fig:teaser}(a4)) for users to search for content of interest. The system embeds users' queries and retrieves them from conversation history, marking related nodes with red circles in the Brush View to help users locate topics of interest (\textbf{T4}).

In Brush View and Content View, the alignment of topics on the y-axis is critical for effective visualization. Therefore, Wiggle Sort is employed to optimize the positions of topics on the y-axis, aiming to reduce the vertical movement distance for users when perceiving visual elements in conversation visualization (\textbf{T1}). 

Wiggle Sort\cite{zhu2022vac2,sun2023application,liu2013storyflow,zhu2022towards,tanahashi2012design} is an approach that employs mathematical optimization models to determine the optimal positions for events along each row. Wiggle distance serves as a classic aesthetic metric in timeline visualization techniques, aiming to minimize the perceived movement distance in a particular direction when users observe visual elements within the timeline. Reducing wiggle distance can, to a certain extent, render a more compact distribution of visualization elements. Consequently, we utilize the Wiggle Sort method to establish the topics' positions on the y-axis.

The existing topic relationship can be essentially represented as a graph with a fixed number of nodes, denoted by $G=(V,E)$. The adjacency matrix of the graph can be written as $A=(a_{ij})$. $a_{ij}$ denotes the number of transitions from topic $i$ to topic $j$ during the conversation. The positions of these topics are then coded in a discrete manner from top to bottom, and 0-1 decision variables are introduced $x_{ij}$, where:
\begin{equation}
x_{ij} = \begin{cases} 1 & \text{when the } i\text{-th topic is at the } j\text{-th position} \\ 0 & \text{when the } i\text{-th topic is not at the } j\text{-th position} \end{cases}
\end{equation}
Using the above decision variables, a mathematical optimization model of the following form is constructed in this study:
\begin{equation}
\label{min}
\min \sum_{i, j, k, l}|j-l| x_{i j} x_{k l} a_{i k}
\end{equation}
\begin{equation}
    \text{s.t.} 
    \left\{
    \begin{array}{ll}
        \sum_{i=1}^{n} x_{ij} = 1 & \forall j=1\ldots n \\[2ex]
        \sum_{j=1}^{n} x_{ij} = 1 & \forall i=1\ldots n \\[2ex]
        x_{ij} = \{0,1\} & \forall i,j=1\ldots n
    \end{array}
    \right.
    \label{constrains}
\end{equation}
The constraints in Eq.(\ref{constrains}) are assignment constraints, which ensure that each topic can only be at one location and there can only be one topic at each location. As the variables in the equations are discrete, the objective function (Eq.(\ref{min})) is a quadratic function, and the constraints are linear, the model can be classified as an Integer Quadratic Programming model. In fact, the problem presented is a typical Quadratic Assignment Problem (QAP\cite{loiola2007survey,fujiwara2017visual}). QAP is considered to be NP-hard, however, considering the relatively small number of topics in the conversation history, this study employs the Gurobi solver to address the aforementioned problem.

\subsection{Topic View}
\vspace{-0.5em}
\begin{figure}[htbp]
  \centering
  \vspace{-0.5em}
  \includegraphics[width=\columnwidth]{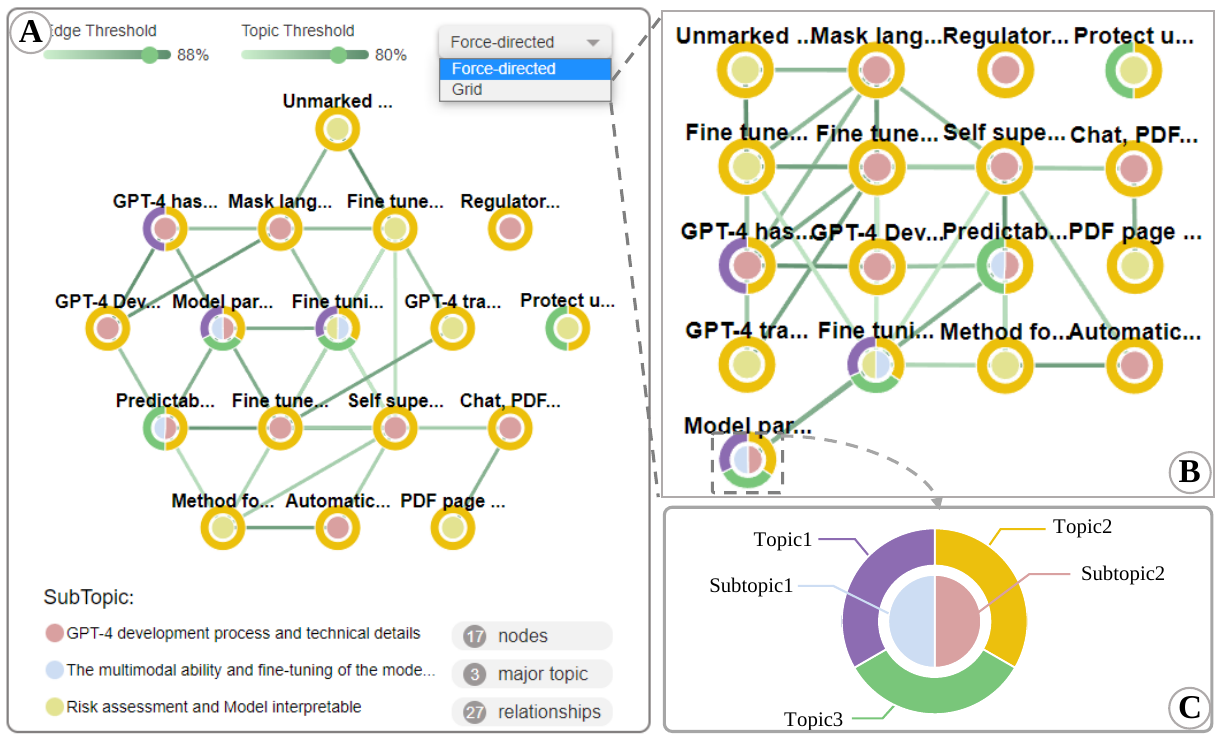}
  \vspace{-1em}
  \caption{Topic View. (A) Uniform distribution of force-directed layout. (B) Orthogonal grid layout. (C) Nested circular charts representing similarity relationships between nodes and topics.}
  \label{fig:circle}
\end{figure}
In Global View, when users find a topic of interest, they can be redirected to Topic View (Fig. \ref{fig:circle}(A)) by clicking on the topic legend in (Fig. \ref{fig:teaser}(a)) (\textbf{T3}). The Topic View displays all Q\&A nodes and their relationships under the selected topic using the structure of a knowledge graph, further refining subtopics to more effectively organize and present information within the topic. This reveals the knowledge connections between all the nodes in the topic, helping users build a complete knowledge system and review the conversation history within the topic (\textbf{T3}).

Similar to Global View, each node in Topic View represents a round of conversation. \Revised{ When the mouse hovers over each node, a overview diagram is also displayed to provide an overview of that node.Fig. \ref{fig:casestudy}(d2,E) }Unlike in Global View, where only the highest similarity topic is used to encode nodes, this view uses nested circular charts to encode the nodes, as shown in (Fig. \ref{fig:circle}(C)). Each nested circular chart has two layers of circular structures: the outer circular chart and the inner circular chart. The outer circular chart encodes the similarity relationships between the node and different topics. Different topics are represented by sectors coded in different colors (consistent with the color of the topic in Global View), and the area of the sector is proportional to the similarity, i.e., the more similar a node is to the topic, the larger the area of the corresponding sector. This presentation shows the degree of relevance between the current node and various topics, enabling a better understanding of the focus of the conversation. The inner circular chart focuses on showing the similarity relationships between the node and different subtopics, using a similar sector area size and color encoding strategy to present the degree of relevance between the current node and various subtopics, further analyzing the details of the conversation.

In the Topic View, the similarity between interconnected nodes is indicated by the color and thickness of the solid lines connecting the nodes, where higher similarity corresponds to thicker lines and darker colors. To reduce visual burden and help users explore relationships within the topic, we allow users to adjust the similarity threshold to filter nodes with higher similarity relationships, improving the readability and intuitiveness of the data. 

To support users in choosing different exploration modes, we provide two optional layout methods (\textbf{T3}): uniform distribution of force-directed layout (Fig. \ref{fig:circle}(A)) and orthogonal grid layout (Fig. \ref{fig:circle}(B)). \textbf{The force-directed graph layout} aims to minimize edge crossings and node overlaps by automatically optimizing node positions. In this layout strategy, we set forces in both the x-axis and y-axis directions to achieve uniform distribution of nodes within a two-dimensional space. At the same time, we introduce repulsive forces between nodes to avoid overlapping between nodes, specifically represented as: 
\begin{equation}
\vspace{-0.5em}
    f_i^{rep} = \sum_{j \neq i}^{n} k \frac{(r^2)}{d_{ij}^2}(p_i - p_j)
    \vspace{-0.2em}
\end{equation}
where constant $k$ indicates a scaling factor, and $r$ denotes the constant representing the desired separation between nodes. The term $d_{ij}^2$ denotes the squared Euclidean distance between nodes $i$ and $j$, while $p_i - p_j$ represents the vector from node $i$ to node $j$.
Furthermore, we set the tension of connecting lines to be proportional to the similarity between nodes, bringing similar nodes closer together, specifically represented as: 
\begin{equation}
\vspace{-0.5em}
    f_i^{attr} = \sum_{j \neq i}^{n} k s_{ij} d_{ij}(p_i - p_j) 
    \vspace{-0.2em}
\end{equation}
where $s_{ij}$ represents the similarity between node $i$ and node $j$. Finally, we set the center of the force-directed graph layout to be in the center of the view to achieve a better visualization. This layout method supports users in exploring related conversation history starting from nodes of interest. \textbf{The grid layout}\cite{kieffer2015hola} divides the two-dimensional space into a regular grid and orders the nodes from left to right and top to bottom based on temporal relationships. In addition, based on nodes' importance, degree, and associated subtopic, we can dynamically adjust node positions to meet specific visual needs. The layout strategy emphasizes the temporal relationship and visual neatness among nodes, which facilitates users' understanding of the internal evolution of topics and the association among nodes.
\subsection{Context-associated Q\&A View}

To fully showcase the detailed information of a specific node and provide relevant contextual information to enhance contextual continuity for ChatGPT (\textbf{T5}), we have designed a Context-associated Q\&A View (Fig. \ref{fig:teaser}(c)), which consists of three visual panels: the Q\&A View (Fig. \ref{fig:teaser}(c1)), the Context View (Fig. \ref{fig:teaser}(c2)), and the Asking View (Fig. \ref{fig:teaser}(c3)).

When exploring and discovering an concerned node in the Global View or Topic View, users can click on the node to view the relevant conversation information. Then, users will be guided to the Q\&A View, which displays detailed Q\&A information of the node. To reduce the cognitive burden of users in reading large amounts of text, the system highlights important content by highlighting keywords or sentences, allowing users to quickly locate and understand the key information (\textbf{T4}). When users intentionally ask new questions, they can enter the questions in the Asking View. As the questions are entered, the Brush View section of the Global View is highlighted with nodes that have a higher similarity to the question. If some nodes are located before the ``forgotten line'' of the Brush View, users can decide whether to include them in the context list to alleviate the context forgetting problem encountered by the model. This will enhance the relevance of information and thus strengthen the coherence of multi-round conversations and the accuracy of answers. Once the nodes are added to the context list, they will be displayed as a nested circular chart in the Context View to facilitate users in tracking the real-time contextual information of the issue of concern. After the user has submitted the question and the relevant contextual information, the Asking View will display an accurate and relevant answer generated by ChatGPT.

In summary, in the Context-associated Q\&A View, we have designed three visual panels to effectively showcase the detailed information of a specific node and allow users to select contextual nodes to provide rich contextual information, thus addressing the context forgetting problem that arises when dealing with multi-turn conversations and enhancing contextual continuity for ChatGPT.

%% file: sections/Evaluation.tex
\section{Evaluation}
\subsection{Case Study}

We have invited a graduate student named Jim to use our system, whose research focus is on information visualization and visual analysis. Jim  plans to use ChatGPT to read the GPT-4 technical report published by OpenAI. Given the length of the report, he needs to go through multi-turn conversations to get all the required information. To ensure a coherent and relevant reading process, he decides to complete the reading within the same session. However, after multiple rounds of conversations with ChatGPT, Jim gradually loses control of the entire conversation, such as forgetting the content of the conversation. Moreover, the model fails to effectively relate to earlier parts of the conversation when answering questions.

$\bullet$ \textbf{Stage 1: Initial Exploration of Global View}

\begin{figure*}[!t]
  \centering
  \includegraphics[width=\textwidth]{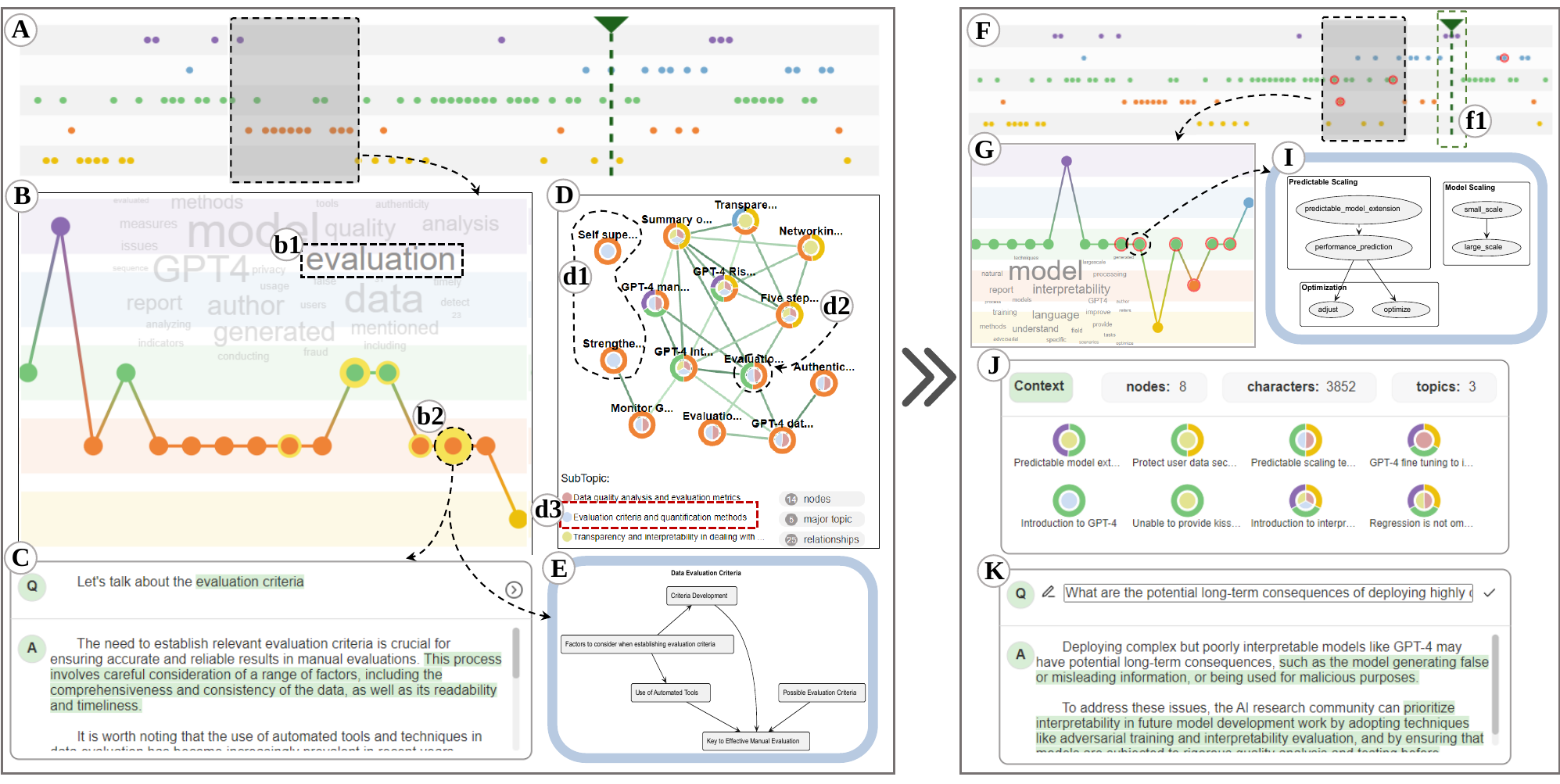}
  \vspace{-2em}
  
  \caption{Jim's process of exploring conversation history and posing questions using C\textsuperscript{5}. On the left side, (A) users brush the region on the Brush View corresponding to the cluster of nodes with the topic \textit{Risk assessment}; (B) the conversation nodes in the selected region are displayed; (C) the detailed information of the selected conversation nodes; (D) the knowledge graph related to the topic of \textit{Risk assessment}; (E) the overview diagram of the node when hovering the mouse over it. On the right side, (F) the highlighted conversation nodes on Brush View after posing a question; (G) the conversation nodes in the brushed region; (I) the overview diagram of a specific node; (J) the context list; (K) the questions and answers.}
  \vspace{-1.5em}
  
  \label{fig:casestudy}
\end{figure*}

After Jim imported his conversation history with ChatGPT to C\textsuperscript{5}, he first focuses on the Global View (Fig. \ref{fig:teaser}(a)). In this view, he notices that the historical conversation content is categorized according to topics as \textit{model development process}, \textit{Risk assessment}, \textit{model interpretability}, \textit{future application prospects}, and \textit{model capability expansion and enhancement}, represented in different colors and presented in the bottom left corner of the Global View. By analyzing the Brush View (Fig. \ref{fig:casestudy}(A)), he is able to observe the distribution of each topic on the timeline, he finds that the historical conversations mainly revolved around \textit{Model interpretability} (colored with green), with a period of focused discussion on \textit{Risk assessment} (colored with red).

$\bullet$ \textbf{Stage 2: Analyzing the Content View and Brush View}

To dive into the details of the discussion on \textit{Risk assessment}, Jim brushes the region in the Brush View (Fig. \ref{fig:casestudy}(A)) corresponding to the cluster of red nodes, and the Content View (Fig. \ref{fig:casestudy}(B)) zooms in to show the nodes of the conversation in the brushed region. By observing the word cloud in the Content View, he learns the keywords of the conversation and is captivated by the information related to ``evaluation'' (Fig. \ref{fig:casestudy}(b1)). When the mouse hovers over the word, the conversation nodes are highlighted at different levels depending on the word frequency to provide hints to users.

$\bullet$ \textbf{Stage 3: Exploring Conversation Nodes in detail}

Jim is interested in the node $n_{36}$ (Fig. \ref{fig:casestudy}(b2)) with the highest level of highlighting and hovers over it to see an overview diagram (Fig. \ref{fig:casestudy}(E)) of the node's conversation content. Next, he clicks on the node to view the detailed conversation content on Q\&A View (Fig. \ref{fig:casestudy}(C)). In the process, he notices a new keyword ``evaluation criteria'' (highlighted in the Q\&A view) and types it into the search box to find highly relevant global conversation nodes in Brush View, thus exploring their content.

$\bullet$ \textbf{Stage 4: Investigating the Topic View}

After an initial exploration of the Global View, he wants to further investigate the content about \textit{Risk assessment}, and clicks on the corresponding legend in Global View. In the Topic View (Fig. \ref{fig:casestudy}(D)), a knowledge graph of all conversation nodes under this topic is displayed, and the topic is further divided into three subtopics. In this view, $n_{36}$ is displayed in Fig. \ref{fig:casestudy}(d2). By observing the color of the inner circular chart of the node, he found that it was related to two sub-topics, among which was the ``Evaluation criteria and quantification methods'' (Fig. \ref{fig:casestudy}(d3)) sub-topic he was interested in, so he explores the nodes in the knowledge graph that are closely related to this subtopic (Fig. \ref{fig:casestudy}(d1)). Additionally, $n_{36}$ is closely related to six other conversation nodes, so he also reviews their details to further revisit the other conversations. Furthermore, by observing the outer circular color of the node, he discovers that the node is also related to the topic \textit{Model interpretability} and then undertakes a similar exploration process in this topic.

$\bullet$ \textbf{Stage 5: Gaining Context and Posing New Questions}

After approximately 5 minutes of exploration, Jim gains a general understanding of the overall context of the conversation history. By using the knowledge graph in Fig. \ref{fig:casestudy}(D), he conducts in-depth exploration on each subtopic to discover potential connections between nodes. He generates new insights about the topic \textit{Risk assessment} and \textit{model interpretability}, and is prepared to pose the following question in Fig. \ref{fig:casestudy}(K): ``What are the possible long-term consequences of deploying highly complex but poorly interpretable models like GPT-4, and how can the field of AI research address these issues in future model development by prioritizing interpretability?''

$\bullet$ \textbf{Stage 6: Submitting questions and Receiving Relevant Answers}

After posing the question, the BrushView highlights conversation nodes that are highly relevant (Fig. \ref{fig:casestudy}(F)) to this question. \Revised{He focuses on the highlighted nodes (Fig. \ref{fig:casestudy}(G)) located before the ``forgotten line'' (Fig. \ref{fig:casestudy}(f1)), browses through their overview diagram (Fig. \ref{fig:casestudy}(I)), and clicks on those that pique his interest to view the full conversation content.} He then adds some of the historical conversations to the contextual list (Fig. \ref{fig:casestudy}(J)). Then, by clicking the Submit button, the system submits the question along with the contextual information to ChatGPT, the answer will be displayed in Fig. \ref{fig:casestudy}(K).

$\bullet$ \textbf{Stage 7: Real-time Updates and Improved Contextual Associations of CharGPT}

The system then fetches the content of this Q\&A and updates all views in real time after text processing, enabling ChatGPT to correlate the content of conversations at any time, improving the performance of contextual associations in multi-turn conversation scenarios.

\Revised{
We record Jim's time investment and analyzed the quality of his interactions. Initially, during his first interaction with the system, Jim spends 3 minutes exploring the conversation history.  Notably, this duration is approximately half of the time it would have taken him to browse through the dialogue history individually without the aid of this system. After posing his first question, he takes an additional 37 seconds to review the associated conversation history and add them to the context. In subsequent interactions, specifically regarding the topic of ``model interpretability'', he no longer needs to extensively explore the conversation history, but only adds relevant historical dialogues to each query, which take him an average of 30 seconds per query.

To further substantiate this analysis, we have included a selection of questions along with their answers both with and without the addition of context in the supplementary materials. These supplementary materials provide insights into how the addition of context impacts the quality and relevance of the responses.

In addition, in the user study section under UT2, we invited more users to rate the answers from ChatGPT with and without the addition of context. The results from this evaluation indicate that the inclusion of context in our approach does indeed improve the quality of the responses.
}

Throughout these stages, Jim successfully recovers a full comprehension of conversations by using the C\textsuperscript{5} for exploration. This process demonstrates the effectiveness and reliability of the system, further validating its value as a tool for analyzing and understanding conversation history and establishing contextual associations of ChatGPT.

\subsection{User Study}

To assess the utility and effectiveness of C\textsuperscript{5}, we conducted a user study. The study aimed to evaluate whether ChatGPT users can use our system to have quality conversations with ChatGPT, observe their exploration process, reflect on future improvements, and collect feedback about C\textsuperscript{5}.

  \textbf{Participant}: We recruit 8 ChatGPT Plus users (5 males, 3 females) to implement a controlled study. The participants varied in age from 20 to 28, including 2 doctoral students, 2 master students and 4 undergraduate students from the Department of Computer Science at a university. All users have a high level of experience with ChatGPT, with usage times ranging from 2 to 5 hours per day. The research interests of the users are primarily focused on the field of visualization.

  \textbf{Task}: We invite users to complete two tasks, followed by collecting their feedback during system usage.

\underline{UT1}: To ensure a fair distribution of expertise and experience, each group is assigned one doctoral student, one master's student, and two undergraduate students. They are instructed to explore the conversation history between Jim and ChatGPT. The first group utilizes the ChatGPT web page, while the second group uses our system. The exploration time is limited to 20 minutes. During this process, participants are required to complete a questionnaire comprising five objective questions (each worth two points) and one subjective question (worth ten points). The specific content of the questionnaire is provided in the supplementary  material.

\underline{UT2}: After completing UT1, all eight participants are asked to pose three questions about the conversation. We will compile and summarize five representative questions, providing two answers for each: one generated with the question and context in ChatGPT, and the other generated with only the question. Participants are required to rate the quality of both answers separately using a five-point Likert scale without knowing the source.

  \textbf{Procedure}: The study begins with a 10-minute introduction, explaining the research objectives and the system's pipeline. Participants are then allowed to freely explore our system for 15 minutes to familiarize themselves with its operation, and we encourage them to ask questions during the exploration process. After completing the training, we ask participants to complete the above tasks and collect the corresponding results for analysis. Finally, the session ended with a semi-structured interview (15 min) and a post-study questionnaire (5-Point Likert Scale). Each session was run in-lab using a 27-inch monitor, following a think-aloud protocol.
\subsubsection{Results}
Regarding \textbf{UT1}, the group utilizing the system for exploration achieved significantly higher average objective (mean value $\mu$ = 8, standard deviation $\sigma$ = 1.41) and subjective scores ($\mu$ = 8, $\sigma$ = 0.82) compared to the group using the web page (objective: $\mu$ = 5.5, $\sigma$ = 1.00; subjective: $\mu$ = 4.75, $\sigma$ = 0.96). \Revised{Additionally, the group using our system took an average of 8 minutes to complete the tasks, whereas the group browsing directly took approximately 19 minutes. The results indicate that C\textsuperscript{5} can not only improve users' comprehension of the conversation history but also reduce the time needed for exploration.}



\begin{figure}[htbp]
  \centering
  \includegraphics[width=0.9\columnwidth]{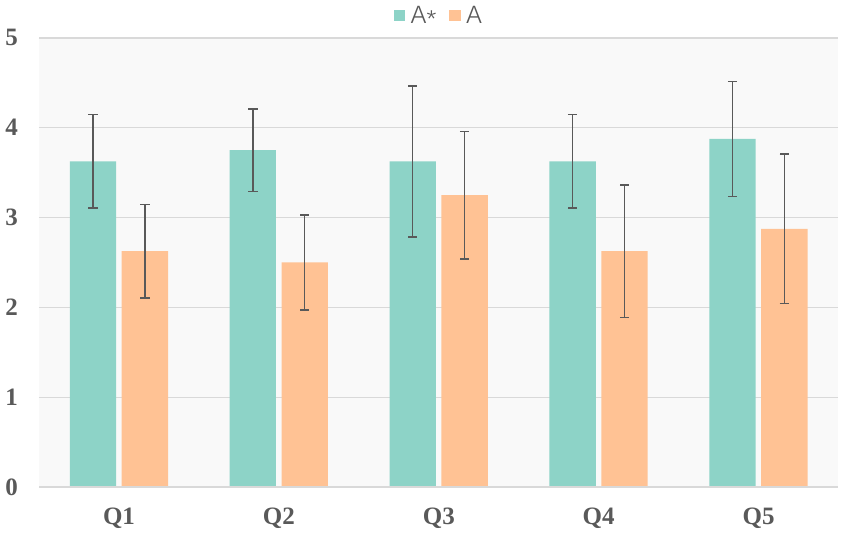}
  \vspace{-1.5em}
  \caption{Comparison of participant rating results in UT2: $\mu$ and $\sigma$. Participants provided ratings for answering specific context-enhanced questions (A*) and for answering questions without additional context (A).}
  \vspace{-1.5em}
  \label{fig:user study}
\end{figure}

Regarding \textbf{UT2}, the results are presented in Fig. \ref{fig:user study}. It demonstrates that, for each question, participants consistently rated answers more favorably when the specific context information was incorporated ($\mu$ = 3.75, $\sigma$ = 0.64). The scores of answers without additional context were significantly lower ($\mu$ = 2.75, $\sigma$ = 0.76). This outcome suggests that providing specific context information improves the performance of ChatGPT in generating more relevant and accurate responses, which is exactly what our system is designed to do. 

In conclusion, the analysis of both tasks highlights the effectiveness of the system in enhancing user comprehension and emphasizes the importance of incorporating specific context information for ChatGPT. The users' feedback is summarized as follows:

\textbf{Usefulness}: Participants in our user study evaluated the usefulness of our interactive conversation visualization system, C\textsuperscript{5}. The system achieved high scores for usefulness ($\mu$ = 4.25), with all participants confirming that the system significantly improves their ability to review historical conversations. The system's usefulness stems from its intuitiveness and efficiency (U1-U8), allowing users to easily explore conversation history and obtain contextual information, ultimately improving the efficiency of multi-turn conversations. Participants generally noted that C\textsuperscript{5} offers new insights and approaches for multi-turn conversation scenarios. Overall, the feedback from the study highlights the usefulness of our system in enhancing the effectiveness of ChatGPT in multi-turn conversation scenarios.

\textbf{Effectiveness}: The results of user tasks indicate that C\textsuperscript{5} significantly improves users' comprehension of conversation history, as reflected in the higher objective and subjective scores achieved by participants who used our system for exploration compared to those who explored the original content of the conversations on the ChatGPT web page. Besides, C\textsuperscript{5} achieved high scores for effectiveness ($\mu$ = 4).  These findings suggest that our system provides an effective solution to the challenges of human forgetting and model contextual forgetting in multi-turn conversation scenarios, and it has the potential to enhance the efficiency of discussions. In the post-study questionnaire, users also provided positive feedback on the system's effectiveness, citing its intuitive design and ability to help them efficiently explore and grasp the content of conversations.

\textbf{Satisfaction}: User feedback regarding C\textsuperscript{5} demonstrated high satisfaction ($\mu$ = 4.125). Participants appreciated the intuitive visual elements and clear layouts, which made it easier to understand the conversation content and structure. They also valued the ability to provide context-specific information when posing questions, which improved the efficiency of discussions in multi-turn conversation scenarios. Some users suggested that additional features could be added to further improve the system, such as allowing users to add tags or categories to conversation nodes for future review and analysis. Overall, the positive user satisfaction rating suggests that the system is effective in addressing the challenges of human forgetting and model contextual forgetting, and has the potential for further improvement.
\subsubsection{Observations, Feedback, and Future Opportunities}
  \textbf{Adaptive Topic Granularity}: Some participants expressed the desire for the system to adaptively adjust the granularity of topics based on the user's focus or interest. For instance, in certain scenarios, users may want to dive deeper into specific topics or sub-topics to gain a more thorough understanding of the conversation. This feedback suggests that an adaptive topic granularity mechanism could be incorporated into the system to enable users to explore conversations at different levels of detail. Developing such a mechanism would require further research in natural language processing and machine learning to effectively identify and represent different levels of topic granularity.

\textbf{Enhanced Context-aware Assistance}: Users appreciated the ability to provide context-specific information when posing questions, which improved the efficiency of discussions in multi-turn conversation scenarios. However, some participants suggested that the system could  automatically recommend or suggest relevant contextual information based on the user's input or browsing history. This would reduce the cognitive load on the user and further improve the conversation experience. To achieve this, future research could focus on developing context-aware recommendation algorithms that can accurately identify and suggest relevant contextual information to the user.

\textbf{Customizable Node Tagging}: Several participants expressed interest in being able to add tags or categories to conversation nodes, which would facilitate reviewing and analyzing conversation history. This feature would allow users to better organize and categorize specific parts of the conversation according to their preferences and needs. Moreover, the tags could also serve as visual cues or reminders, enabling users to quickly locate and access relevant content within the conversation. Future research can explore designing and implementing a flexible and user-friendly tagging feature that integrates with the existing visualization interface, enabling users to effectively manage and review their multi-turn conversations.

%% file: sections/DISCUSSION_AND_LIMITATIONS.tex
\section{DISCUSSION AND LIMITATIONS}
Although the evaluation results show the effectiveness of the system and methodology, we believe there are other considerations that could further improve the techniques used in this work. In this section, we present the lessons learned from this work, the limitations of the system, and potential future research directions.

  \textbf{Generalizability.} During the design process of our visualization system, we considered its adaptability to different types of multi-turn conversation scenarios. By applying topic mining and multi-level text data generation methods, we are able to handle various types of dialogue data. Meanwhile, the global view, topic view and context-associated Q\&A view in the system provide users with multiple levels of analysis and exploration. This makes our approach highly generalizable and can be applied to other multi-turn conversation scenarios, such as online forums, email exchanges, etc.

  \textbf{Lessons learned.} We present two lessons learned during the development of C\textsuperscript{5}.

\underline{Data Visualization Selection}: Selecting an appropriate visualization scheme according to data characteristics. To effectively present the conversation history between users and ChatGPT, we processed the historical conversation data, classifying it by topic, resulting in a data structure containing different sets of topics. Some conversations may belong to multiple topics, leading to overlaps among these sets. Initially, we tried the classic Venn diagram\cite{cui2021vinemap} to visualize the global view, but it was insufficient for representing complex relationships. We then chose the metro map\cite{jacobsen2020metrosets}, which has advantages in representing complex relationships and local clustering. However, we found limitations such as poor support for collection data with global temporal information, low spatial utilization, and slow optimization speed when the data volume is large.

After conducting further research, we have discovered that GitLog diagrams are better suited for conversation scenarios. They are particularly effective in displaying data that has chronological and branching characteristics, enabling users to track and comprehend the progression of conversations more easily. Moreover, GitLog diagrams optimize space usage and facilitate exploration of locally significant features. Therefore, we have chosen to use a design inspired by GitLog diagrams to represent the conversation history more effectively.

\underline{LLM-Visualization Synergy}: Exploring the synergy between LLMs and visualization technology. LLMs bring many positive impacts to the visualization field, such as their excellent semantic understanding ability to automatically recognize and extract key information within the text. In our work, we have used LLMs like GPT and observed that they simplify the workflow and achieve better topic modeling results. Compared to LDA models, LLMs do not require a predetermined number of topics, thus avoiding subjectivity.

Visualization technology also helps address the challenges of LLMs. In our work, we used visual design to engage users in improving the model's performance in context association. Specifically, we use visualization to direct users' attention to potential contextual information. When posing questions, we enhance the model's performance in conversation comprehension and generation by drawing the user's attention to possible contextual information through specific visual designs and allowing the user to include it in the contextual list. This synergy provides new insights and practical experience for visualization research.

  \textbf{Limitations and future work.} Two limitations are observed in the proposed system. First, the visualization scheme in this study primarily focuses on the global perspective and local details of the conversation history, but an in-depth investigation into the interaction process between users and ChatGPT has not been conducted. Therefore, future research could consider utilizing visualization tools to explore interaction patterns and strategies between users and ChatGPT. \Revised{For instance, linguistic analysis might provide another avenue to explore the structure of dialogues. We are pondering over different approaches, such as incorporating more granular linguistic patterns or employing more advanced visualization techniques to analyze and represent the interaction patterns and strategies between users and ChatGPT.}Secondly, the visualization approach presented in this paper is suitable for small to medium-scale conversation data, while large-scale conversation data may necessitate more efficient algorithms and scalable visualization designs. Consequently, future work could explore adopting multi-level and multi-scale representations to display conversation data, enabling users to explore and analyze the data at different levels and scales. In addition to these limitations, we also identified several opportunities for future work, including adaptive topic granularity, enhanced context-aware assistance, customizable node tagging, and combined exploration of multiple conversation histories. By addressing these limitations and incorporating the suggested future work, C\textsuperscript{5} can be further refined to better accommodate users' diverse needs in multi-turn conversation scenarios with ChatGPT.

%% file: sections/Conclusion.tex
\vspace{-0.2em}
\section{Conclusion}
In this paper, we presented C\textsuperscript{5}, a web-based visual analytic system designed to analyze the conversation history between users and ChatGPT, thereby enhancing users' conversation comprehension and ensuring contextual continuity for ChatGPT. Our work effectively solve the problem of human forgetting and model contextual forgetting. C\textsuperscript{5} first employs a text embedding and topic modeling approach based on GPT to divide the conversation history into s hierarchical topic-based text data with temporal information. Then, all information becomes accessible through the three interactive views that compose the system. Finally, we demonstrate the usability and effectiveness of our system through one case study and one user study.

%% file: sections/Acknowledgment.tex
\section{ACKNOWLEDGMENT}
This work is supported by Zhejiang Provincial Natural Science Foundation of China (LR23F020003, LTGG23F020005) and 
National Natural Science Foundation of China (61972356, 62036009). Furthermore, thanks to the participants for
taking part in the requirement analysis and evaluation. Guodao Sun is the corresponding author.